\pdfoutput=1

\documentclass[11pt]{article}

\usepackage[]{ACL2023}

\usepackage{times}
\usepackage{latexsym}

\usepackage[T1]{fontenc}

\usepackage[utf8]{inputenc}

\usepackage{microtype}

\usepackage{inconsolata}
\usepackage{booktabs}

\usepackage{multirow}
\usepackage{adjustbox}
\usepackage{amsmath}
\usepackage{color}

\usepackage{listings}

\definecolor{dkgreen}{rgb}{0,0.6,0}
\definecolor{gray}{rgb}{0.5,0.5,0.5}
\definecolor{mauve}{rgb}{0.58,0,0.82}

\def\Section {\S}

\newcommand{\squishlist}{
\begin{list}{$\bullet$}
{ \setlength{\itemsep}{0pt}
	\setlength{\parsep}{3pt}
	\setlength{\topsep}{3pt}
	\setlength{\partopsep}{0pt}
	\setlength{\leftmargin}{1.5em}
	\setlength{\labelwidth}{1em}
	\setlength{\labelsep}{0.5em} } }

\newcounter{Lcount}
\newcommand{\squishlisttwo}{
\begin{list}{\arabic{Lcount}. }
	{ \usecounter{Lcount}
		\setlength{\itemsep}{0pt}
		\setlength{\parsep}{0pt}
		\setlength{\topsep}{0pt}
		\setlength{\partopsep}{0pt}
		\setlength{\leftmargin}{2em}
		\setlength{\labelwidth}{1.5em}
		\setlength{\labelsep}{0.5em} } }

\newcommand{\squishend}{\end{list} }

\title{Leveraging Training Data in Few-Shot Prompting for Numerical Reasoning}

\author{Zhanming Jie \\
  ByteDance Research\\
  \texttt{allan@bytedance.com} \\\And
  Wei Lu \\
  StatNLP Research Group \\
  Singapore University of Technology and Design \\
  \texttt{luwei@sutd.edu.sg} \\}

\begin{document}
\maketitle
\begin{abstract}
Chain-of-thought (CoT) prompting with large language models has proven effective in numerous natural language processing tasks, but designing prompts that generalize well to diverse problem types can be challenging~\cite{zhou2022least}, especially in the context of math word problem (MWP) solving.
Additionally, it is common to have a large amount of training data that have a better diversity coverage but CoT annotations are not available, which limits the use of supervised learning techniques. 
To address these issues, we investigate two approaches to leverage the training data in a few-shot prompting scenario: \textit{dynamic program prompting} and \textit{program distillation}.
Our approach is largely inspired by \citet{gao2022pal}, where they proposed to replace the CoT with the programs as the intermediate reasoning step. 
Such a prompting strategy allows us to accurately verify the answer correctness through program execution in MWP solving.
Our dynamic program prompting involves annotating the training data by sampling correct programs from a large language model, while program distillation involves adapting a smaller model to the program-annotated training data.
Our experiments on three standard MWP datasets demonstrate the effectiveness of these approaches, yielding significant improvements over previous baselines for prompting and fine-tuning.
Our results suggest that leveraging a large amount of training data can improve the generalization ability of prompts and boost the performance of fine-tuned small models in MWP solving\footnote{Our code and data are available at \url{https://github.com/allanj/dynamic-pal}.}.

\end{abstract}

\section{Introduction}
Designing effective prompts is crucial for the success of few-shot prompting with large language models (LLMs) in tasks requiring complex reasoning skills~\cite{wei2022chain,zhou2022least,shrivastava2022repository,fu2022complexity}. 
Especially for the task of arithmetic word problem, it poses a significant challenge to design  a small number of  chain-of-thought (CoT) prompts~\cite{wei2022chain} to solve a wide range of  problems.

\begin{figure}[t!]
	\centering
	\adjustbox{max width=1.0\linewidth}{
		\includegraphics[]{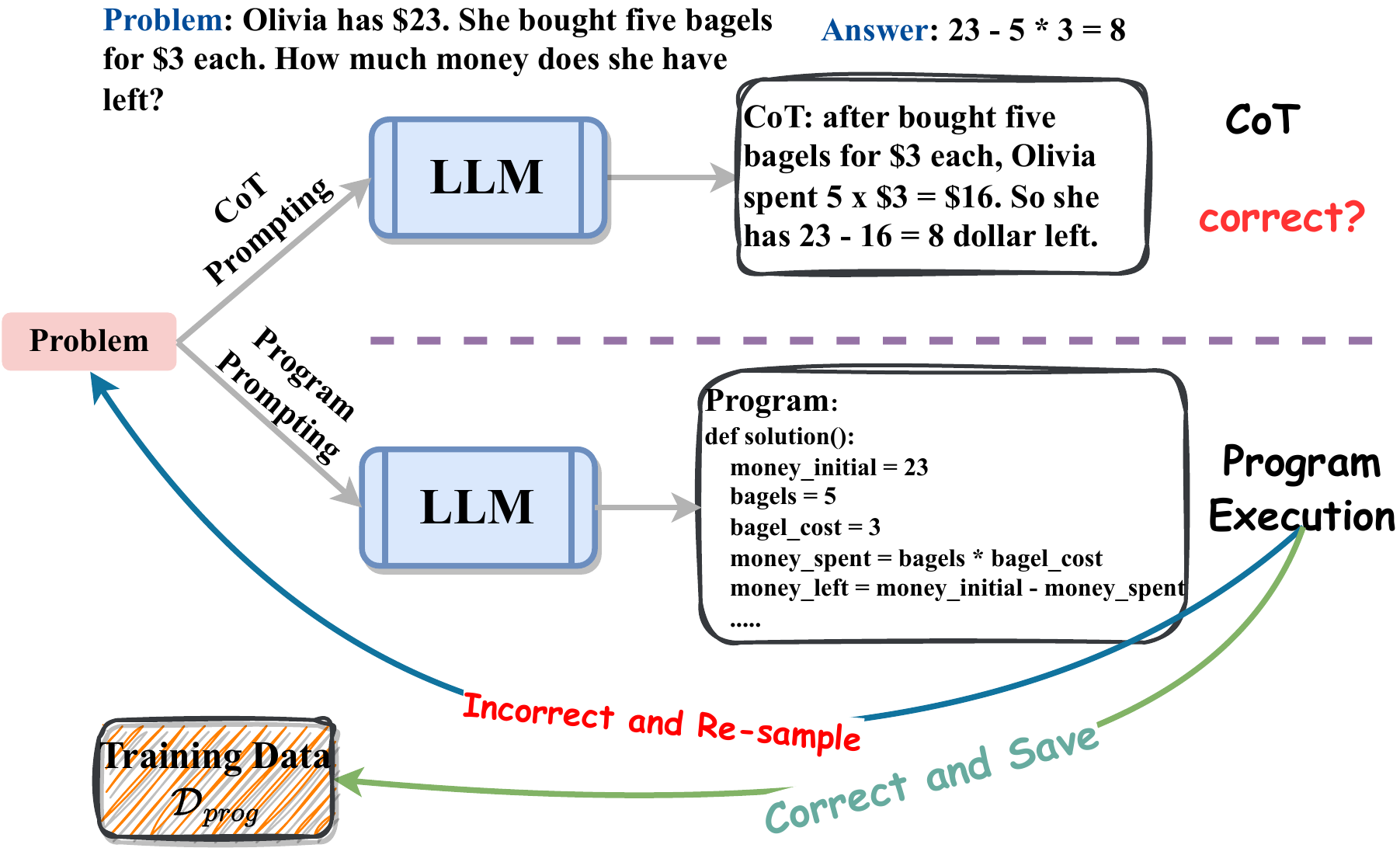}
	}
	\caption{Program annotation with LLM.}
	\label{fig:introduction}
\end{figure}

Fortunately, a modest amount of training data is usually available though no chain-of-thought (CoT) annotation exists.
\citet{rubin2021learning} adopts a retrieval-based approach to select similar samples as prompts. 
While such a method does not work well for numerical reasoning compared to CoT prompting~\cite{wei2022chain}, recent work~\cite{magister2022teaching} also tried to distill the CoT knowledge from large language models to smaller language models. 
The distilled CoT annotations allow us to further fine-tune the small language models.
However, there is no guarantee that the generated CoT prompts for the training data are correct, and it is challenging to perform an automatic evaluation to verify the CoT correctness. 
As an alternative, recent work~\cite{drori2022neural,gao2022pal,mishra2022lila} has adopted programs as intermediate reasoning chains in tasks such as math word problems (MWPs), allowing for automatic verification of answers through program execution.
Inspired by these approaches, we can perform prompting with code generation models, such as Codex~\cite{chen2021evaluating}, to annotate the training data with programs that can be executed.
Figure \ref{fig:introduction} shows the process of automatic program annotation using large language models. 
As we can see in this example, though the final answer ``\textit{8 dollar}'' by CoT is correctly generated, the intermediate reasoning path is wrong because of incorrect calculation for ``$5\times 3$''. 
Instead, the program sampling is relatively more rigorous in that we can execute to obtain the numeric answer rather than CoT in natural language. 
Apparently, we can keep sampling the program with different temperatures until the answer executed from the program matches the correct one. 
Once we obtain the annotated program, we can use the ``\textit{annotated}'' training data with the pseudo-gold program to further improve the performance on the test set.

\begin{figure}[t!]
	\centering
	\begin{lstlisting}[language=Python]
def solution():
"""Natalia sold clips to 48 of her friends in April, and then she sold half as many clips in May. How many clips did Natalia sell altogether in April and May?"""
clips_april = 48
clips_may = clips_april / 2
clips_total = clips_april + clips_may
result = clips_total
return result
	\end{lstlisting}
 \vspace*{-3mm}
	\caption{Example program from the GSM8K training set following the format in PAL.}
	\label{fig:program}
\end{figure}

In this work, we primarily study two approaches for making use of the ``\textit{annotated}'' programs: \textit{dynamic program prompting} and \textit{program distillation}~\cite{magister2022teaching}. 
Dynamic program prompting employs the top-$k$ similar training samples (with annotated pseudo-gold programs) as few-shot prompts. 
We use publicly available and state-of-the-art sentence encoders such as OpenAI embeddings~\cite{neelakantan2022text}\footnote{\url{https://openai.com/blog/new-and-improved-embedding-model/}} and Sentence-T5~\cite{ni2022sentence} for computing the cosine similarity. 
On the other hand, we follow \citet{magister2022teaching} to fine-tune smaller language models on our pseudo-gold training data. 
Overall, our experiments on three standard math word problem datasets demonstrate the effectiveness of leveraging the training program in our few-shot prompting.
We observe significant improvements for all datasets, especially for the MathQA~\cite{amini2019mathqa} dataset, where diverse subjects (e.g., physics, probability, etc.) were involved in the problems where the fixed prompt  accompanied by limited examples is insufficient to encapsulate the entire scope of requisite knowledge.


\section{Approach}
\label{sec:approach}
\paragraph{Training Data Annotation}
Following the approach in program-aided language model (PAL)~\cite{gao2022pal}, we can sample the program for each math word problem as an annotation. 
Specifically, we use the math prompts from PAL as seed prompts to perform few-shot prompting with large language models (i.e., Codex~\cite{chen2021evaluating}). 
We follow the exact same format from PAL~\cite{gao2022pal} without any changes, Figure \ref{fig:program} shows an example program from the GSM8K training set.
We can verify the answer's correctness by comparing the result from the program execution with the ground-truth value.

For each math word problem $\boldsymbol{x}$ in training set $\mathcal{D}$, we first perform greedy decoding with temperature $T = 0$ to obtain the bet Python program. 
If the predicted answer $\hat{y}$ from the executed program $\boldsymbol{P}$ matches the ground-truth answer $y$, we add this tuple $(\boldsymbol{x}, \boldsymbol{P}, y)$ into a new training data set $\mathcal{D}_{prog}$. 
If the predicted answer is incorrect, we increase the temperature and continue sampling programs until we find one with the correct answer. 
In practice, we may not always obtain the correct answer and have a limited budget for Codex API usage. 
Thus, we sample at most $K$ times for each instance.
If we cannot find a program with the correct answer within $K$ samples, we discard the instance $\boldsymbol{x}$. As a result, the size of the resulting training set $\mathcal{D}_{prog}$ is expected to be smaller than the original training set (refer to Table \ref{tab:dataset}).


\begin{table}[t!]
	\centering
	
	\adjustbox{max width=1.0\linewidth}{
		\begin{tabular}{lcccc}
			\toprule
			\multirow{1}{*}{\textbf{Dataset}} &  \multirow{1}{*}{\#\textbf{Train}} &\textbf{\#Program} & \multirow{1}{*}{\#\textbf{Valid}} & \multirow{1}{*}{\#\textbf{Test}}  \\
			\midrule
			GSM8K & \textcolor{white}{0}7,473 & 6,363 (85.1\%) & - & 1,319 \\ 
			SVAMP &  \textcolor{white}{0}3,138 & 3,071 (97.9\%)& - & 1,000\\
			MathQA$\dagger$ & 16,191 & 7,676 (47.4\%) & 2,411 & 1,605 \\
			\bottomrule
		\end{tabular}
	}
	\caption{Dataset statistics and the percentage of annotated programs. $\dagger$: We follow \citet{jie2022learning} to obtain the preprocessed split.}
	\label{tab:dataset}
\end{table}
\subsection{Dynamic Program Prompting}

\begin{table*}[t!]
	\centering
	\adjustbox{max width=1.0\linewidth}{
		\begin{tabular}{clcccc}
			\toprule
			& \textbf{Model} & \textbf{\#Param}& \textbf{GSM8K} & \textbf{SVAMP} & \textbf{MathQA} \\
			\midrule
			\multirow{8}{*}{Prompting}&LaMDA~\cite{thoppilan2022lamda} & 137B & 17.1 & - & -\\
			&PaLM~\cite{chowdhery2022palm} & 540B & 58.1 & 79.0 & - \\
			&GPT-3 CoT (\texttt{\textcolor{darkblue}{text-davinci-002}})& 175B& 48.1 &- & - \\
			&Codex CoT (\texttt{\textcolor{darkblue}{code-davinci-002}})& 175B & 65.6 & 74.8& 29.9 \\
			&Complex CoT~\cite{fu2022complexity} & 175B & 55.4 & - & 36.0$\dagger$ \\
			&PAL~\cite{gao2022pal} & 175B& 72.0& 79.4&  -  \\
			&PAL (\texttt{reproduced}) & 175B& 71.6& 77.4& 30.0 \\
			\cmidrule[1pt]{2-6}
			&Our Dynamic Program Prompting & 175B &\textbf{76.6} &\textbf{80.3} &\textbf{61.7} \\
			\midrule\midrule
			\multirow{5}{*}{Fine-tuning}& GPT-3~ & 175B & 33.1 & - &- \\
			& CoT Fine-tune~\cite{magister2022teaching} & \textcolor{white}{0}11B & 38.2& - & -  \\
			& CoT Fine-tune (CodeGen) & \textcolor{white}{00}6B & 35.3&  40.2 & 25.3 \\
			\cmidrule[1pt]{2-6}
			& Our Program Distillation & \textcolor{white}{00}6B & \textbf{39.0} &  \textbf{48.0} & \textbf{50.6} \\
			\bottomrule
		\end{tabular}
	}
	\caption{Performance comparison over previous approaches using prompting and fine-tuning. $\dagger$: not directly comparable as they use less amount of test data. }
	\label{tab:mainresult}
\end{table*}

\paragraph{Prompt Retrieval}
Given all the instances $(\boldsymbol{x}, \boldsymbol{P}, y)$ in $\mathcal{D}_{prog}$, we retrieve the top $M$ most relevant instances as prompts. 
We use state-of-the-art sentence embeddings such as sentence-T5~\cite{ni2022sentence} and SimCSE~\cite{gao2021simcse} to obtain the representation for each math word problem $\boldsymbol{x}$. 
We then compute the cosine similarity between each test sample and all 
training samples. 
Based on the similarities, we select the most similar $M$ exemplars from the training instances in $\mathcal{D}_{prog}$.

\paragraph{Similarity}
To further verify the effectiveness of using similarity to select the prompting exemplars, we also experiment with alternative strategies such as random selection from $\mathcal{D}_{prog}$ and selecting the exemplar with the least similarity.

\subsection{Program Distillation}
Our purpose is to train a smaller model using the annotated data compared to LLMs such as Codex~\cite{chen2021evaluating}.  
In order to do this, we follow the approach of fine-tuning a pre-trained model on $\mathcal{D}_{prog}$, similar to \citet{magister2022teaching}.
Given a math word problem $\boldsymbol{x}$, our objective is to generate the corresponding Python program $\boldsymbol{P}$. 
We use the publicly available CodeGen~\cite{nijkamp2022codegen} for this task as it is trained specifically for code generation and pre-trained models are available\footnote{\url{https://huggingface.co/Salesforce/codegen-16B-mono}}.
CodeGen is a standard Transformer-based~\cite{vaswani2017attention} autoregressive model.

\section{Experiments}

\paragraph{Dataset and Experiment Setting}
Similar to \citet{fu2022complexity}, we mainly conduct experiments on GSM8K~\cite{cobbe2021training}, SVAMP~\cite{patel2021nlp}, and MathQA~\cite{amini2019mathqa} datasets. 
Table \ref{tab:dataset} shows the statistics and the number of annotated programs. 
The programs are annotated via few-shot prompting with PAL~\cite{gao2022pal} (\Section \ref{sec:approach}). 
We perform prompting with Codex (\texttt{code-davinci-002}) where the API usage is free. 
Following \citet{gao2022pal}, we set the maximum token for the generation to $600$. 
The training set in the SVAMP dataset is the easiest as we can obtain pseudo-gold programs about 98\% 
We only managed to obtain program annotations for 47.4\% of the instances for MathQA, as it is the most challenging and noisy-labeled~\cite{fu2022complexity} with diverse problem types (e.g., physics, probability, geometry, etc).

The maximum number of sampling $K$ for each training instance is set to 5\footnote{We chose $K$ = $5$ to strike a balance between cost and efficiency. Increasing $K$ may not lead to significant improvements.}, and the temperature $T$ is $0.5$ following previous practice~\cite{zelikman2022star}.
We discard the training instance if we cannot find a proper program. 
The number of prompts $M$ is set to 8 following previous work in math word problem solving~\cite{gao2022pal,fu2022complexity,wei2022chain}. 
In fine-tuning experiments, we use the 6B CodeGen language model.
The learning rate for fine-tuning experiments is $2$e-$5$. 
We fine-tune the CodeGen model with a batch size of $48$ and experiment with $40$ epochs on all datasets. 
The fine-tuning Experiments are conducted with $8$ A100 GPUs. 
We did not perform a hyper-parameter search for fine-tuning. 
All parameters are set to the above default values.

\begin{table}[t!]
	\centering
		\setlength{\tabcolsep}{4pt} 
		\renewcommand{\arraystretch}{1.1} 
	\adjustbox{max width=1.0\linewidth}{
			\begin{tabular}{llccc}
					\toprule
					 & & \textbf{GSM8K} & \textbf{SVAMP} & \textbf{MathQA} \\
					\midrule
					\multirow{2}{*}{Most Similar} & OpenAI & 76.6&	80.3&	61.7 \\
					\multirow{2}{*}{$M$ Exemplars}& \texttt{SimCSE}~\cite{gao2021simcse} & 76.4 & 80.1 & 61.0\\
					& \texttt{ST5}~\cite{ni2022sentence}  &76.6&	79.9	&61.6\\
					\midrule
					Random & - & 74.4 & 78.1 & 34.0 \\
					\midrule
					\multirow{2}{*}{Least Similar} & OpenAI &  73.5 & 78.2 & 34.1 \\
					\multirow{2}{*}{$M$ Exemplars}& \texttt{SimCSE}~\cite{gao2021simcse} & 76.0 & 78.4 & 34.7\\
					& \texttt{ST5}~\cite{ni2022sentence}  &74.2 & 77.9&  34.3\\
					\bottomrule
				\end{tabular}
		}
	\caption{Performance comparison among different sentence representations.}
	\label{tab:similarity}
\end{table}

\paragraph{Main Results}
We conduct both prompting and fine-tuning experiments on all datasets.  
Table \ref{tab:mainresult} shows the performance comparison with previous prompting approaches using large language models. 
Similar to PAL~\cite{gao2022pal}, our results demonstrate that program-based approaches achieve the best performance across all the datasets. 
Our approach, based on the same underlying mechanism as PAL, achieves new state-of-the-art performance (at the time of submission) with a 175B model and obtains at most $5$-point absolute improvements over PAL. 
On the easiest dataset, SVAMP, we still achieve a $0.9$-point improvement over the best-performing baseline and $2.9$ points better than the reproduced PAL. 
On the MathQA dataset, known for its noise, we see significant improvements of over $20$ points compared to other prompting baselines. 
The observed substantial enhancement suggests that the utilization of in-context examples facilitates the model's comprehension of the questions and enables it to generate solutions based on analogous prompts.
These results suggest that retrieving similar examples is crucial, especially for  complex datasets.

In addition to our prompting approach, we also evaluate the effectiveness of fine-tuning a smaller language model on the annotated training data, as shown in Table \ref{tab:mainresult} (bottom section). 
We fine-tune a 6B CodeGen model on the training data with annotated programs, and our approach achieves better performance with $0.8$-point improvement than an 11B T5 model on the GSM8K dataset. 
We use the same method as \citet{magister2022teaching} to perform prompting on the training set and obtain the annotated CoT. 
Notably, for the SVAMP dataset, our fine-tuning approach with programs significantly outperforms fine-tuning with natural language CoT by $7.8$ points. 
On the MathQA dataset, which is known to have noisy labels, our fine-tuning performance is significantly better than vanilla prompting performance. 
The dynamic program prompting achieves over 30-point improvements compared with PAL. 
Compared with CoT Fine-tuning approach using CodeGen, our program distillation approach is also $25.3$ points better in accuracy.
This observation further highlights the importance of leveraging the training data in complex datasets. 
In general, the fine-tuning performance with smaller models is worse than few-shot prompting with large language models on GSM8K and SVAMP, indicating that program distillation may not be sufficient to compensate for the generalization limitations of smaller language models.


\begin{figure}[t!]
	\centering
	\adjustbox{max width=1\linewidth}{
		\includegraphics[]{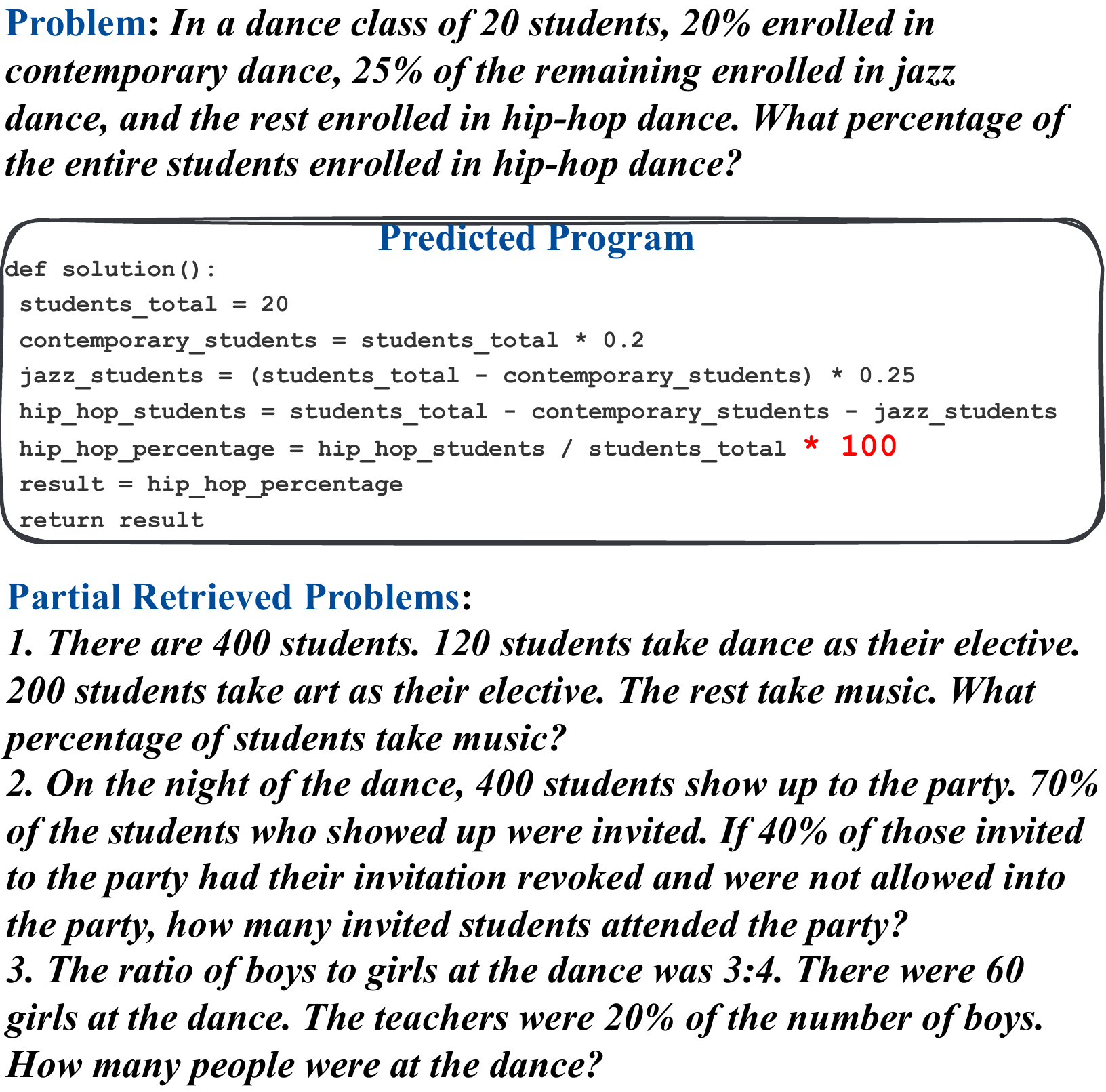}
	}
	\caption{Example prediction by our prompting approach and the corresponding retrieved problems.}
	\label{fig:analysis}
\end{figure}

\paragraph{Prompt Retrieval Strategy}
To further justify the effectiveness of using the most similar exemplars as prompts, we conduct experiments with different prompt retrieval strategies and the results are presented in Table \ref{tab:similarity}. 
``\textit{Random}'' strategy is to randomly select $M$ exemplars as the prompt.
The table shows that using different sentence embeddings results in consistent performance when using the ``\textit{most similar $M$ Exemplar}'' strategy. 
However, using the ``\textit{least similar exemplars}'' consistently leads to a drop in performance, especially on the MathQA dataset where the evaluation data is more similar to the training data~\cite{fu2022complexity}.
Moreover, the least similar exemplars are unlikely to encompass the full spectrum of information required in the MathQA dataset where a broader range of knowledge exists.
The ``\textit{Random}'' strategy also shows similar performance as using the ``\textit{least similar exemplars}'', indicating that neither of them provides additional benefits compared to using the  ``\textit{most similar exemplars}'' strategy.


\paragraph{Qualitative Prompt Analysis}

To gain insights into how the prompts affect performance, we compare the results between PAL and our approach on the GSM8K dataset. 
The retrieved prompts by our approach have a higher level of word level overlapping with the question. 
Figure \ref{fig:analysis} shows an example of how our approach helps in making more accurate predictions. 
The code ``\texttt{* 100}'' marked in red is the information that PAL failed to generate. 
This suggests that PAL may not have been confident about the  ``\textit{percentage}'' for this question. 
Our prompts, on the other hand, contain many questions related to ``\textit{percentage}'' which are more likely to help the model make correct predictions. 
However, we also note that the similarity-based method is not always better than fixed prompts by PAL. 
On GSM8K, PAL still performs better on $5.5$\% of the questions while our similarity-based approach performs better on $10.3$\% of all questions. 
Thus, similarity-based prompts can produce positive improvements in general.

\section{Related Work}
Our work is mostly related to recent literature that incorporates the training data to improve the language model performance on downstream tasks.
\citet{chung2022scaling} shows that we can benefit from additional CoT data for both large and small language models. 
\citet{li2022advance} samples CoT reasoning paths for the training data and uses them to diversify the prompts on the GSM8K dataset. 
Alternatively, we can use the sampled CoT to further fine-tune the language models~\cite{huang2022large,magister2022teaching,meng2022generating}.
In practice, we cannot guarantee the correctness of the sampled CoT, especially for the task of math word problem solving, which requires rigorous reasoning paths. 
Recent approaches~\cite{magister2022teaching,wang2022list} attempt to reduce the negative effect by matching the answer with generated CoT or assigning different weights for the samples.
Simultaneously with this study, \citet{uesato2022solving} proposes to use step-based reward to improve the performance specifically on GSM8K. 
In order to do so, the authors need to  annotate a portion the data to train the underlying reward model. 
However, these methods cannot completely avoid the underlying limitation as it is challenging the evaluate the step-by-step natural language CoT~\cite{golovneva2022roscoe,prasad2023receval}.
Our approach is inspired by program generation via few-shot prompting~\cite{gao2022pal}, we perform prompting on the training data and easily verify the answer correctness by executing the program, which allows us to obtain more reliable pseudo-gold programs.

\section{Conclusion and Future Work}

Motivated by program-based prompting~\cite{gao2022pal,drori2022neural}, we are able to obtain the pseudo-gold program as the intermediate reasoning step for training data. 
We then present two approaches to make use of such data with program annotations in both of the few-shot prompting and fine-tuning scenarios.
In few-shot prompting with LLMs, we sample similar exemplars as prompts for experiments. 
In the fine-tuning approach, we directly fine-tune a pre-trained language model on program-annotated data.
Our experiments demonstrate both few-shot prompting and fine-tuning can significantly benefit from the training data annotated with programs, especially for complex problems in the MathQA dataset. 

For future research, our goal is to design a structured model that leverages the potential of data with program annotations, particularly in light of the substantial underperformance of smaller language models. Interestingly, even with their limitations, structured models~\cite{jie2022learning,shao2022chaining} have exhibited the capacity to outshine large language model prompting on MathQA. 
Additionally, the recent emergence of instruction-following models~\cite{ouyang2022training,wang2022self}, exemplified by Alpaca~\cite{alpaca}, has prompted our interest in equipping large language models with mathematical reasoning capacities~\cite{wang2023msat} while maintaining the integrity of their underlying language understanding capabilities.


\section*{Limitations}
The methods we have employed for prompting and fine-tuning have yielded noticeable improvements, yet certain limitations persist within practical applications. 
To achieve optimal performance, we continue to rely on prompting using large language models, which prove to be costly for the research community.
Furthermore, retrieval efficiency may present a challenge when dealing with extensive training sets, as identifying the top $M$ exemplars for each example becomes increasingly time-consuming.
Consequently, devising a more efficient algorithm to expedite the retrieval process represents a potential area for future exploration.

Despite the potential for performance improvement by sampling 40 reasoning paths for each question as presented by \citet{wang2022self,fu2022complexity}, we were unable to incorporate this approach due to budget constraints.
Additionally, although training data has proven beneficial, the gains for smaller models are insufficient to surpass the performance of large language models. 
This observation may indicate the necessity for a fundamentally different model design or a superior pre-trained model (e.g., Galactica~\cite{taylor2022galactica} or Code-T5~\cite{wang2023codet5+}) as a more effective basis for fine-tuning.

\section*{Acknowledgement}

We would like to thank the anonymous reviewers, our meta-reviewer, and senior area chairs for their constructive comments and support on our work.
This research/project is supported by the National Research Foundation Singapore and DSO National Laboratories under the AI Singapore Program (AISG Award No: AISG2-RP-2020-016), and the Ministry of Education, Singapore, under its Tier 3 Programme (The Award No.: MOET32020-0004).

\bibliography{custom}
\bibliographystyle{acl_natbib}




%



\end{document}